# X-Ray CT Reconstruction of Additively Manufactured Parts using 2.5D Deep Learning MBIR

Amirkoushyar Ziabari[*,1], Michael Kirka[2], Vincent Paquit[1], Philip Bingham[1], and Singanallur Venkatakrishnan[1]

[1.] Imaging, Signals and Machine Learning Group, Oak Ridge National Lab
[2.] Deposition Science and Technology, Oak Ridge National Lab
* Corresponding author: aziabari@ornl.gov

Rapid non-destructive evaluation (NDE) of additively manufactured (AM) parts is critical to advancing our understanding of the impact of various process parameters and qualifying the final quality of the built parts. X-ray computed tomography (XCT) is an important non-destructive technique that can image parts in 3D at the micro-meter resolution. XCT involves taking X-ray images of the part from different orientations and computationally processing the images to obtain a 3D volumetric image. However, obtaining high-resolution 3D reconstructions is typically time consuming, as it requires a large number of images to be taken. This presents a fundamental roadblock to the adoption of the technology for rapid inspection of a large number of parts that may be produced on a given day.

We have been developing novel algorithms [1-3], termed model-based iterative reconstruction (MBIR) techniques, based on Bayesian estimation techniques to obtain high quality 3D reconstructions for CT from significantly fewer and noisy measurements. We showed that compared to the standard methods such as the filtered back projection (FBP), MBIR can dramatically reduce the measurement time (by 32x) [2] and in turn accelerate the scan time of a single part. However, using MBIR for real-time high-quality reconstruction is challenging as, due to its iterative nature, it requires a significant amount of computation.

In this paper, we present a deep learning algorithm to rapidly obtain high quality CT reconstructions for AM parts. In particular, we propose to use CAD models of the parts that are to be manufactured, introduce typical defects and simulate XCT measurements. These simulated measurements were processed using FBP (computationally simple but result in noisy images) and the MBIR technique. We then train a 2.5D deep convolutional neural network [4], deemed 2.5D Deep Learning MBIR (2.5D DL-MBIR), on these pairs of noisy and high-quality 3D volumes to learn a fast, non-linear mapping function. The 2.5D DL-MBIR reconstructs a 3D volume in a 2.5D scheme where each slice is reconstructed from multiple inputs slices of the FBP input. Given this trained system, we can take a small set of measurements on an actual part, process it using a combination of FBP followed by 2.5D DL-MBIR. Both steps can be rapidly performed using GPUs, resulting in a real-time algorithm that achieves the high-quality of MBIR as fast as standard techniques. Intuitively, since CAD models are typically available for parts to be manufactured, this provides a strong constraint "*prior*" which can be leveraged to improve the reconstruction.

We voxelized a CAD model of an AM part into a volume of 512 slices of 256×256, and added random defects, mainly voids and holes of different sizes, to the volume. This is shown in Fig. 1a. We simulated a parallel beam XCT measurement with Poisson statistic, where we used 180 views, each correspond to one degree rotation. An example projection of the AM part is shown in Fig. 1c. Next, we performed reconstruction of noisy projections using FBP and MBIR methods. We divided the reconstructed volume to two sub-volumes of 256×256×256. The first volume was used for training the 2.5D DL-MBIR network. Each set of 256×256×5 neighboring slices (with a stride of 1 slice) in 3D volume was divided into nine

Research sponsored by the U.S. Department of Energy, Office of Energy Efficiency and Renewable Energy, Advanced Manufacturing Office, under contract DE-AC05-00OR22725 with UT-Battelle, LLC.
This manuscript has been authored by UT-Battelle, LLC, under contract DE-AC05-00OR22725 with the US Department of Energy (DOE). The US government retains and the publisher, by accepting the article for publication, acknowledges that the US government retains a nonexclusive, paid-up, irrevocable, worldwide license to publish or reproduce the published form of this manuscript, or allow others to do so, for US government purposes. DOE will provide public access to these results of federally sponsored research in accordance with the DOE Public Access Plan (http://energy.gov/downloads/doe-public-access-plan).

patches of 128×128×5. We used data augmentation to increase the number of training data. The trained network was then used to perform reconstruction on the second sub-volume of data. Promising results were obtained which validated the method. Therefore, we performed a full reconstruction of the volume from a different perspective that wasn't seen by the 2.5D DL-MBIR network. Fig. 1b shows the reconstructed volume using 2.5D DL-MBIR. We compared the results with reconstructions obtained by FBP and MBIR methods in Fig. 1d-g. Three cross sections along the center of the 3D volume are plotted. The Peak-Signal-to-Noise-Ratio (PSNR) value for each reconstructed volume was computed with respect to the ground truth volume and are listed in Table 1. These results demonstrate that both MBIR and 2.5D DL-MBIR outperform FBP in terms of image quality. In addition, as can be seen in Table 1, the 2.5D DL-MBIR high quality results were obtained at about the same speed as FBP and much faster than MBIR.

|  | FBP | MBIR | 2.5D DL-MBIR |
| --- | --- | --- | --- |
| PSNR (dB) | 24.24 | 53 | 51.23 |
| Computational Time (s) | 5.6 | 3060 | 6.7 |

While the results for the 2.5D DL-MBIR method is promising, the method can be improved in several aspects. First, we need to generalize it, so it can handle different types of defects, noise levels and images that are obtained at different resolutions. In addition, we intend to further enhance the method to enable reconstruction from scans with limited number of views. Moreover, while 2.5D DL-MBIR can achieve the same quality as MBIR, we intend to improve it to obtain better than MBIR reconstructions.

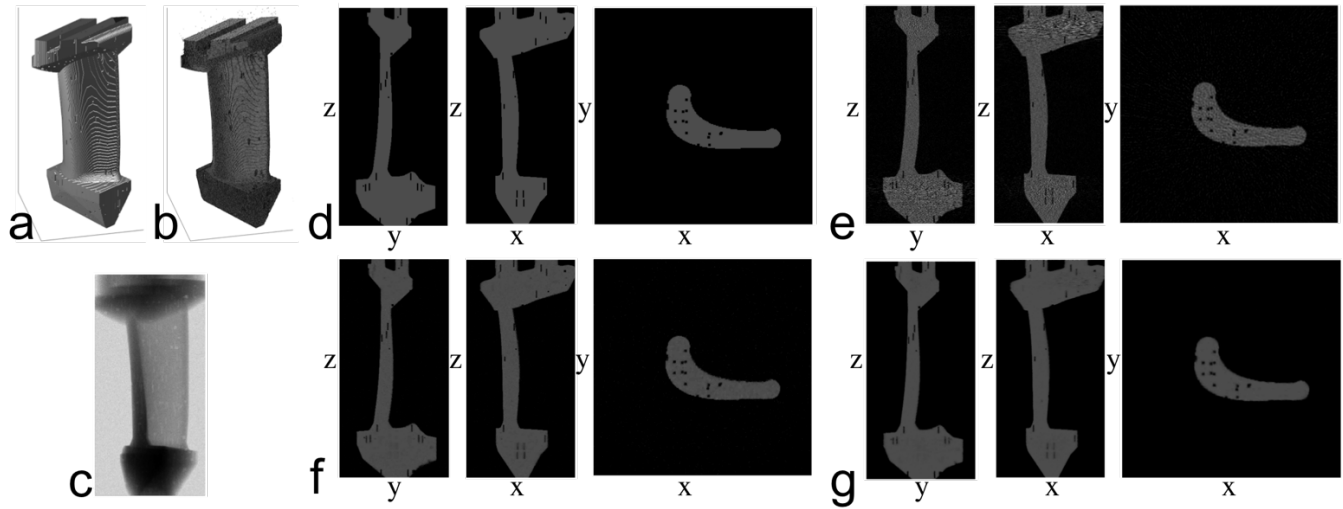

**Figure 1.** a. 3D Volume. b. 2.5D DL-MBIR reconstruction. c. A projection at θ=90°. Comparison between 3 slices along the center of the volume: d. Ground Truth. e. FBP. f. MBIR. g. 2.5D DL-MBIR.